# Development of a High-Resolution Indoor Radon Map Using a New Machine Learning-Based Probabilistic Model and German Radon Survey Data

*Eric Petermann,[1] Peter Bossew,[1*] Joachim Kemski,[2] Valeria Gruber,[3] Nils Suhr,[1] and Bernd Hoffmann[1]*

[1]Section Radon and NORM, Federal Office for Radiation Protection (BfS), Berlin, Germany
[2]Sachverständigenbüro Dr. Kemski, Bonn, Germany
[3]Department for Radon and Radioecology, Austrian Agency for Health and Food Safety, Linz, Austria

**Background:** Radon is a carcinogenic, radioactive gas that can accumulate indoors and is undetected by human senses. Therefore, accurate knowledge of indoor radon concentration is crucial for assessing radon-related health effects or identifying radon-prone areas.

**Objectives:** Indoor radon concentration at the national scale is usually estimated on the basis of extensive measurement campaigns. However, characteristics of the sampled households often differ from the characteristics of the target population owing to the large number of relevant factors that control the indoor radon concentration, such as the availability of geogenic radon or floor level. Furthermore, the sample size usually does not allow estimation with high spatial resolution. We propose a model-based approach that allows a more realistic estimation of indoor radon distribution with a higher spatial resolution than a purely data-based approach.

**Methods:** A multistage modeling approach was used by applying a quantile regression forest that uses environmental and building data as predictors to estimate the probability distribution function of indoor radon for each floor level of each residential building in Germany. Based on the estimated probability distribution function, a probabilistic Monte Carlo sampling technique was applied, enabling the combination and population weighting of floor-level predictions. In this way, the uncertainty of the individual predictions is effectively propagated into the estimate of variability at the aggregated level.

**Results:** The results show an approximate lognormal distribution of indoor radon in dwellings in Germany with an arithmetic mean of $63\,Bq/m^3$, a geometric mean of $41\,Bq/m^3$, and a 95th percentile of $180\,Bq/m^3$. The exceedance probabilities for 100 and $300\,Bq/m^3$ are 12.5% (10.5 million people affected) and 2.2% (1.9 million people affected), respectively. In large cities, individual indoor radon concentration is generally estimated to be lower than in rural areas, which is due to the different distribution of the population on floor levels.

**Discussion:** The advantages of our approach are that is yields *a*) an accurate estimation of indoor radon concentration even if the survey is not fully representative with respect to floor level and radon concentration in soil, and *b*) an estimate of the indoor radon distribution with a much higher spatial resolution than basic descriptive statistics. https://doi.org/10.1289/EHP14171

## Introduction

The radioactive noble gas radon (Rn) is a human carcinogen.[1,2] A causal association between radon exposure and the risk of developing lung cancer has been established[3]; a possible association with other diseases is currently under scientific discussion.[4–6] The adverse health effects of radon are known for uranium (U) miners[7] and have also been shown to be associated with indoor radon exposure at home.[3,8] Radon is one of the most important causes of lung cancer after smoking. Further, smoking and radon reinforce each other in their harmful effects on human health. Gaskin et al.[9] estimated the number of annual deaths due to indoor radon exposure at 266,000 worldwide (for 66 countries; 74% of world's population), making radon responsible for $\sim 3\%$ of all the $\sim 10$ million cancer deaths per year worldwide reported by the International Agency for Research on Cancer in 2020.[10]

In most cases, the main source of indoor radon is the soil and rock under the house, where radon (Rn-222) is generated by the decay of radium (Ra-226), a progeny of uranium (U-238). Radon concentration is measured by its radioactive activity in becquerels (decay processes per second) per meters cubed. As a component of natural radioactivity, radon is present in all soils and rocks but with distinct spatial variation regarding its concentration (e.g., see the *European Atlas of Natural Radiation*[11]). Radon enters houses mainly via advective transport, driven by temperature- and wind-induced pressure differences, via small cracks and fissures in the buildings' foundation.[12,13] Secondary sources of indoor radon are building materials containing elevated radionuclide levels,[14,15] water supply (especially when water is drawn directly from private wells),[16–18] natural gas[16,19] and, in some specific cases, outdoor air.[20,21] The accumulation of radon indoors depends on the exchange with (usually) low radon outdoor air and is controlled by the air tightness of the building and the ventilation rate. In most cases, indoor radon concentration is higher on the lower floors than on the upper floors because the lower floors are closer to the ground and thus closer to the main source of radon.[14,22]

The estimation of regional or national indoor radon concentration is usually achieved by large-scale measurement campaigns. These measurement campaigns have to be representative so that the sample reflects the relevant characteristics of the population (e.g., availability of geogenic radon, distribution of people on different floors, building types) and thus provides an unbiased estimate. Therefore, sampling design should ensure representativeness by, for example, taking random samples from some type of national registry (e.g., postal addresses, buildings, telephone numbers).[23–25] In these cases, sampling density is proportional to the population density and the sample should reflect the spatial distribution of dwellings or people across the country. Subsequently, the measured data are usually aggregated for the spatial scale of interest, and the desired statistical characteristics can be derived. However, even with a population-weighted sampling design, bias may occur because participation is voluntary

*Retired.

Address correspondence to Eric Petermann, Köpenicker Allee 120-130, 10318 Berlin, Germany. Email: epetermann@bfs.de

Supplemental Material is available online (https://doi.org/10.1289/EHP14171).

The authors declare they have no conflicts of interest related to this work to disclose.

Conclusions and opinions are those of the individual authors and do not necessarily reflect the policies or views of EHP Publishing or the National Institute of Environmental Health Sciences.

and people who know that they are exposed to a higher hazard (e.g., people living on lower floors) might be aware of this fact and may be more motivated to participate. Therefore, even with population-weighted sampling, the characteristics of the sample must also be compared with the characteristics of the population with regard to building-related factors. In general, although a key factor governing indoor radon concentration,[14,26–28] the distribution of the population across floors is often not considered,[24,29,30] probably because of a lack of appropriate data. However, some commendable exceptions exist where the distribution of the population across floors has been explicitly considered, such as by Vienneau et al.[31] for Switzerland.

It is important to note that the purpose of many studies that focus on mapping indoor radon is not necessarily to estimate national indoor radon exposure but, rather, to delineate hazard areas. For the latter, a common approach is to define a reference house as done in studies for Austria,[32] Germany,[33] South Korea,[34] and Switzerland[35] or in the *European Atlas of Natural Radiation*.[11] The results of these studies, although reporting estimates of indoor radon values, cannot be easily used for estimating national indoor radon concentration because the predictions were made for a defined situation (e.g., the ground floor) and thus do not reflect the variability of housing conditions within the population.

To overcome the limitations described above, predictive models are widely applied as a complementary tool for concentration estimation by using available information on the relevant variables that govern indoor radon. The advantages of applying a model-based approach are 3-fold: It allows *a*) correction of a potential sampling bias (lack of sample representativeness), *b*) estimation at unmeasured locations is possible, and *c*) a higher spatial resolution if sufficiently highly resolved predictor data are available. Application of a predictive model requires that spatially exhaustive information on relevant factors is available (e.g., a soil radon map, a national building registry). If this information is known not only globally (e.g., at the national scale in this case), but also on a more highly resolved scale (e.g., at state or district level), a predictive model can also be used to estimate at a higher spatial resolution. This approach is very useful because a robust estimation of indoor radon concentration at high spatial resolution based solely on measured data would require an enormous number of measurements, which would drastically increase the financial and logistical effort.

Machine learning models have been increasingly used to model indoor radon in recent years.[31,36–41] These studies consistently report on machine learning models' superior performance compared with previous approaches. However, predictive models—also those based on machine learning—are only able to explain a certain amount of the observed variability due to the absence of relevant information (e.g., the information on air tightness of an individual building, resident behavior with respect to ventilation frequency and intensity) or because predictor data does not sufficiently reflect local characteristics [e.g., outcrop of (unmapped) small-scale geological units]. As a consequence, predictions of indoor radon at the individual level, that is, the building, dwelling, floor, or room level, usually have a large uncertainty that is manifested by a low predictive performance for point estimates. The amount of prediction uncertainty also affects the estimation of the radon distribution of a population because the distribution of predictions (i.e., expected values) have the tendency to be considerably smoothed in comparison with the observed distribution. For example, Vienneau et al.[31] estimated indoor radon concentrations for each resident in Switzerland using a machine learning-based model with consideration of many relevant predictors. Although accurate for estimating the national mean, the approach used will underestimate the true variability of indoor radon concentration in the population because it does not explicitly account for prediction uncertainty. Therefore, the circumstance that many people will be exposed to radon concentrations that substantially deviate from the expected value (i.e., the conditional mean) is not considered, and thus the estimation of quantities, such as the exceedance probability of certain threshold values (e.g., $300\,\mathrm{Bq/m^3}$), are doomed to fail. Consequently, to derive realistic estimates of certain quantiles or exceedance probabilities, propagation of prediction uncertainty is required.

The overall goal of our study was to characterize the distribution of indoor radon concentration in Germany at different administrative levels (country, federal state, district, municipality). We have chosen a model-based approach to represent the variability of environmental and building-related characteristics within the population and to enable the characterization (including exceedance probabilities) of indoor radon concentration at a high spatial resolution.

## Methods

### *Survey Description*

A population-weighted national indoor radon survey was carried out in Germany from 2019 to 2021. We aimed to gather a population-representative sample by determining a target number of participants per district (*Landkreise und kreisfreie Städte*; $n = 401$ districts in 2019) that was proportional to the population size of the respective district (i.e., more measurements took place in districts with more inhabitants), up to a total number of 7,479 households. A commercially available address register (Deutsche Post Direkt) was chosen as a practicable compromise for the sampling. The address register has some limitations in terms of completeness, such as non–up-to-date or missing addresses due to inconsistencies in data usage. Participants were recruited by mail, and the sample was randomly selected from the address register, with no prefiltering in such as terms of building style, building age, and floor. In total, however, only $\sim 1,350$ households were recruited as participants through the random mailing. Therefore, the survey was advertised via (local) media and press releases and through the Federal Office for Radiation Protection (BfS), the initiator of this study, to increase the number of participants, especially in the still underrepresented districts. In this way, the desired sample size was reached with volunteers. The actual number of measurements differed from the target number in some cases, resulting in an over- or underrepresentation of certain districts (Figure S1).

A number of relevant building- and household-related characteristics were documented using a questionnaire that included the following characteristics: type of building, year of construction, building style, number of apartments in the building, number of persons per household, presence of a basement, basement access, connection of the basement to the living area, tightness of the windows, ventilation, thermal renovation, radon remediation measures, floor, type of room, and contact of the room with the ground. The questionnaire also asked whether radon measurements had already been carried out in the household, why the participants took part in the study, and how they found out about the campaign (e.g., by mailing, press). Georeferencing of the building or dwelling was done via the postal address.

Radon measurements were carried out in two habitable rooms (e.g., living room, bedroom, dining room, children's room, workroom) in each dwelling with solid-state nuclear track detectors (SSNTDs) for 1 y (between fall of 2019 and spring of 2021) according to DIN ISO 11665-4. The detectors were sent by post and returned. There were no specific behavioral instructions (e.g., on ventilation behavior) given that the measurement campaign was intended to reflect typical conditions over the course of a year. In total, 7,479 households ($\sim 87.5\%$ of those enrolled) returned the detectors.

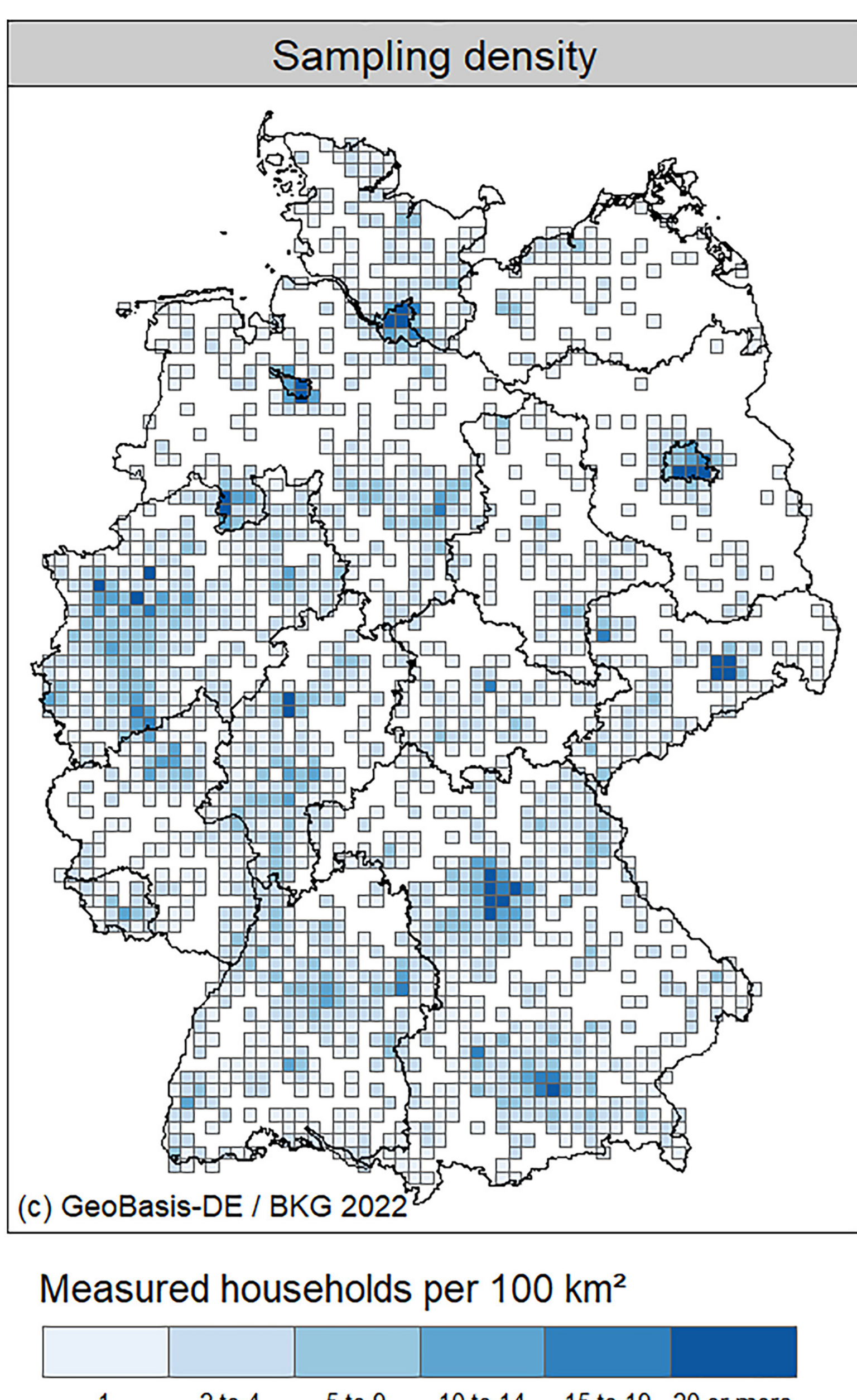

**Figure 1.** Sampling density of households participating in the new national indoor radon survey of Germany, 2019–2021. Sampling density was proportional to the population density at the district level (*Landkreise und kreisfreie Städte*). Consequently, more measurements were taken in urban areas than in rural areas. Generally, two measurements were made in each household ($n = 7{,}479$ households).

The target measurement duration of continuous monitoring for 365 d (with a tolerance of $\pm 10\%$) was achieved by 14,053 detectors. These data (Figure 1) were used for further analysis in this study.

### *Potential Predictor Data*

Numerous studies have shown a clear dependence of indoor radon on geogenic radon availability[14,33,42–48] because geogenic radon is generally the main source of indoor radon. Furthermore, many studies[16,28,38,49] have also proven an association between indoor radon concentration and climate/weather conditions (temperature, precipitation, soil moisture) that result in temporal variability of indoor radon on hourly-to-seasonal timescales. In addition, wind speed has been identified as an influential parameter governing indoor radon by affecting the indoor/outdoor pressure gradient.[49,50] Slope of the terrain has also been found to have some explanatory power for indoor radon modeling.[34,38] Different studies identified tectonic faults as the cause of locally increased geogenic radon.[51,52] Regarding dwelling characteristics, floor level,[14,26–28,53] and building age[14,18,43,45,53,54] have consistently been reported to be main governing factors of indoor radon concentration. Some studies have also identified an influence of building type on indoor radon.[14,16,43,54] The hypothesized association of the model predictors and indoor radon are briefly summarized in the following:

- Building age might be a proxy for air tightness of the foundation, as well as building design (including building material).
- Outdoor radon is another source for radon entry.

- Climate-related predictors [outdoor temperature (in degrees Celsius), annual precipitation (in millimeters)] could be a proxy for building design, ventilation intensity, and living habits.
- The slope inclination (in degrees) affects the contact area with the ground; a larger contact area could enhance radon entry rates.
- Soil gas permeability (in log meters squared) and soil moisture (in percentage usable field capacity) can affect soil radon generation and transport in the ground.
- Wind exposure might be a proxy for typical pressure gradients between the building and atmosphere that govern advective radon influx from soil to building.
- Building type and number of living units affect building design and could be a proxy for living habits and ventilation intensity.
- Geological fault density (in meters per hectare) represents areas where radon transport can be enhanced locally, that is, areas with high fault density may be susceptible to increased geogenic radon supply.
- The household income could be a proxy for building quality and financial resources available for radon remediation.

***Environmental data.*** Environmental predictor data comprised mapped soil characteristics, climate, terrain, and geology (Figure 2). Soil-related characteristics included the soil radon concentration (in becquerels per meter cubed), soil gas permeability (in log meters squared) and soil moisture (in percentage usable field capacity). Soil radon concentration and soil gas permeability reflect the conditions at a 1-m depth and are based on measurements collected by the BfS. Maps were produced at a 500-m resolution using a random forest model. The soil radon map was trained using 10,184 field measurements and 10 predictors: *a*) geology (own classification, data source BGR[55]), *b*) terrestrial gamma dose rate,[56] *c*) soil moisture,[57] *d*) wilting point,[58] *e*) a topographic wetness index [processed via the System for Automated Geoscientific Analyses (SAGA) algorithm[59]], *f*) silt fraction in topsoil,[60] *g*) annual precipitation,[61] *h*) clay fraction in topsoil,[60] *i*) pH of topsoil,[62] and *j*) slope [own calculation; data source: Federal Agency for Cartography and Geodesy (BKG)[63]]. The soil gas permeability map was trained using 9,600 field measurements and 10 predictors: *a*) geology (own classification, for description see above; data source BGR[55]), *b*) silt fraction in top soil,[60] *c*) mean annual outdoor air temperature,[64] *d*) a topographic wetness index (processed via SAGA algorithm[59]), *e*) sand fraction in topsoil,[60] *f*) slope (own calculation; data source: BKG[63]), *g*) potassium in topsoil,[62] *h*) clay fraction in topsoil,[60] *i*) available water capacity[58], and *j*) soil moisture.[57] More details on the generation of predictor data can be found in the Supplemental Material, "S1. Environmental predictor data." Climate data, including soil moisture,[57] annual precipitation[61] (in millimeters), and average outdoor temperature[64] (in degrees Celsius), were taken from the German Weather Service (DWD). Terrain characteristics were calculated based on a digital elevation model at a 25-m resolution.[63] The wind exposure index reflects the relative terrain position and is a dimensionless index with higher values indicating stronger exposure to wind (for details on calculation, see the Supplemental Material, "S1. Environmental predictor data"). The topographic wetness index also describes the relative terrain position and is a measure of water accumulation in the terrain as a consequence of surface water runoff and, thus, is a measure of soil wetness. Tectonic fault density (fault line length in meters per hectare) is based on the faults mapped in the geological map of Germany at a scale of 1:250,000.[55] Radon concentration in outdoor air (in becquerels per meter cubed, at a 1.5-m height) were produced using a random forest model at a 500-m resolution. More details on generating and processing of environmental data can be found in the Supplemental Material, "S1. Environmental predictor data." Summary statistics on assigned environmental data to indoor radon survey data are given in Excel Table S1.

Environmental candidate predictors were as follows:

- Radon concentration in soil (in kilobecquerels per meter cubed)
- Radon concentration outdoors (in becquerels per meter cubed)
- Mean annual outdoor air temperature (in degrees Celsius)
- Mean annual precipitation sum (in millimeters)
- Mean annual soil moisture (in percentage usable field capacity)
- Soil gas permeability (in log meters squared)
- Wind exposure index
- Topographic wetness index
- Slope (in degrees)
- Tectonic fault density (in meters per hectare).

***Building-related data.*** Building-related data were derived from a georeferenced dataset published by the BKG.[65] This dataset refers to the year 2021 and comprises all buildings in Germany ($n = 21{,}866{,}120$). Each building is characterized by a point coordinate. For further analysis, only residential buildings (i.e., buildings with at least 1 inhabitant) were considered ($n = 21{,}343{,}204$). The building dataset contained the following information for each address:

- Number of households
- Number of inhabitants
- Number of (aboveground) floor levels
- Building age (for information on classification, see Supplemental Material, "S2")
- Building type (for information on classification, see Supplemental Material, "S2")
- Net household income.

Owing to some discrepancies between categories defined in the survey vs. BKG datasets, we harmonized the data for building age and type (see Supplemental Material, "S2. Harmonization of building data" for details) to allow predictions for the entire residential building stock. Summary statistics on assigned building data to indoor radon survey data are given in Excel Table S1.

Building-related candidate predictors were as follows:

- Floor level
- Building age
- Number of households
- Type of building
- Net household income (in euros).

***Floor-level distribution of the population.*** The number of people per floor level was estimated using data on the number of floor levels and the number of inhabitants per building. We assumed an equal distribution of people across all floor levels from the ground floor to the uppermost floor level of the building.

Basements, which usually have the highest radon levels within a building owing to their proximity to the ground,[14,22] are widespread in Germany (87% of the measured houses had a full or partial cellar). In a considerable number (19%) of the participating households, measurements were carried out in basements (mostly used as workrooms). However, no external data were available on the occurrence and the use of basements. Therefore, assumptions about the use of basements were necessary to take them sufficiently into account in estimating radon concentration distribution in the population.

Considering that basement occupation depends on the building type, we assumed that basements were more frequently used in townhouses and single- and two-family houses compared with multifamily houses or apartment blocks. Therefore, we parameterized basement occupation as shown in Figure S2: In single- and two-family houses and townhouses, the basement occupation was assumed to be 30% relative to upper floor occupation (factor = 0.3). For all other building types (multifamily house, apartment building, high-rise apartment building, terrace house, farmhouse, and office building) basement occupation was assumed to be 5% relative to upper floor occupation (factor = 0.05).

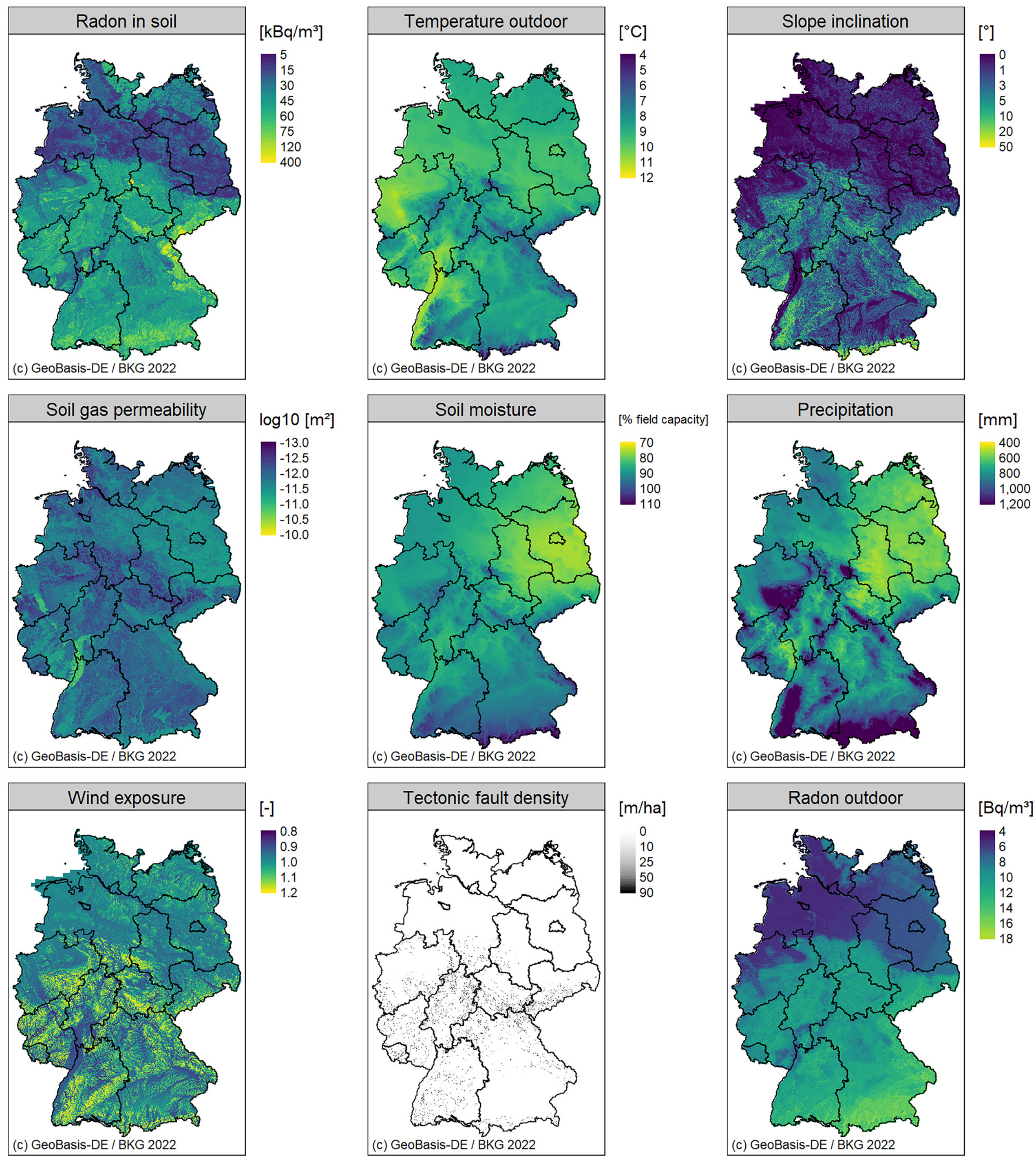

**Figure 2.** Soil characteristics in Germany: radon in soil and soil gas permeability (source: BfS); climate: outdoor temperature, soil moisture, and precipitation (source: DWD); terrain: slope and wind exposure (derived from digital elevation model; source: BKG); geology: tectonic fault density (source: BGR). Data were spatially joined with indoor radon measurements and used as predictors in the quantile regression forest model. Outdoor radon was used for improving the fit of a distributional function fitted to quantile predictions. Note: BfS, Federal Office for Radiation Protection; BGR, Federal Institute for Geosciences and Natural Resources; BKG, Federal Agency for Cartography and Geodesy; DWD, German Weather Service.

***Treatment of missing data.*** Information on building-related characteristics in the BKG dataset was not complete for every building. Data imputation was applied to fill these gaps and to facilitate predictions on the full residential building stock. Missing data concerned information on "number of floor levels" (8% missing), "building year" (1% missing), and "building type" (1% missing).

Missing data on building type was imputed by using information on the number of households. If the number of households was $\leq 2$, then the building type "single- and two-family house" was imputed. If the number of households was >2, then the building type "multifamily house" was imputed.

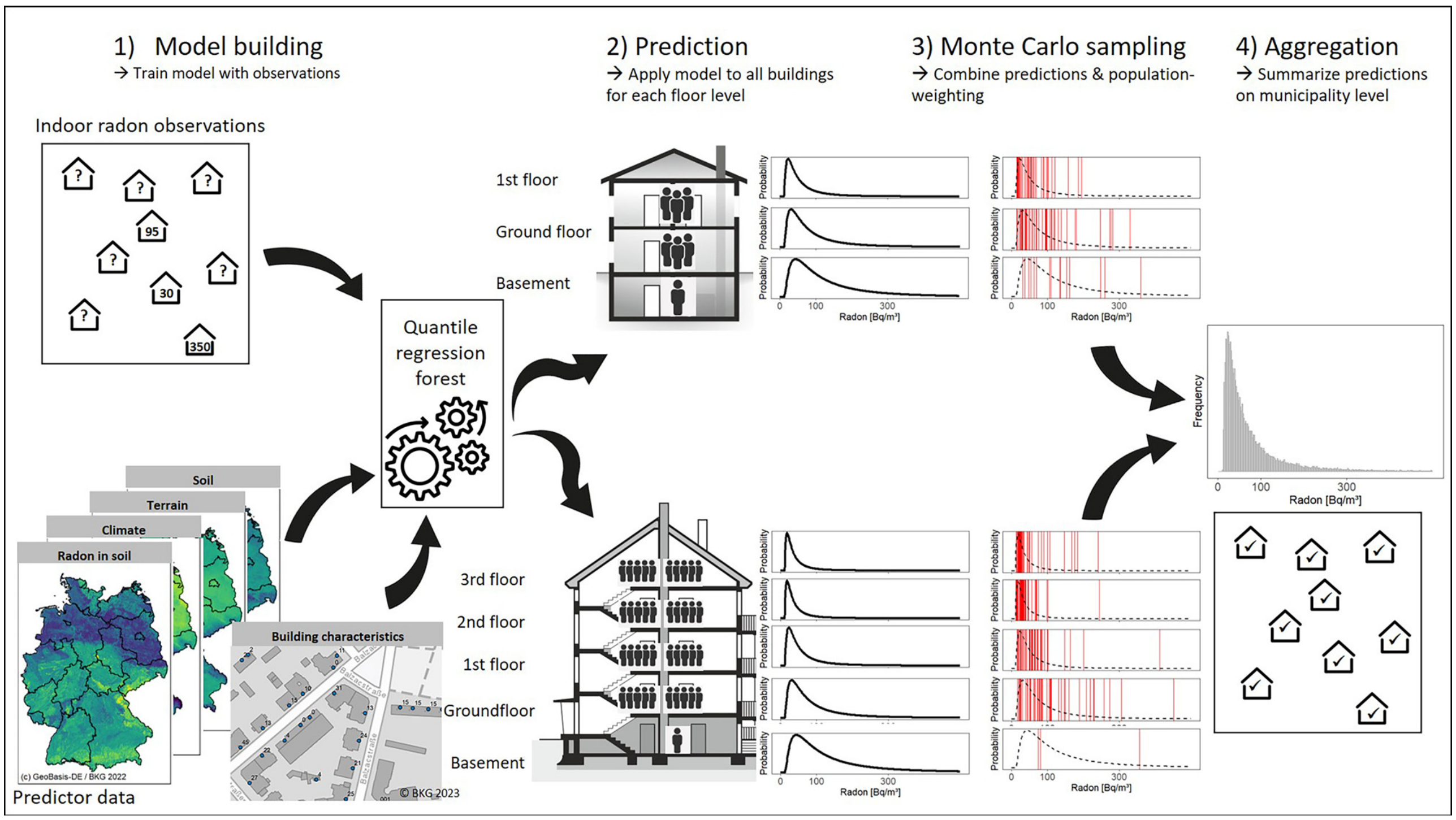

**Figure 3.** Schematic figure of the applied modeling approach: 1) model building: training of the quantile regression forest model using indoor radon observations ($n = 14{,}053$ measurements) and 12 environmental and building-related predictors; 2) prediction of PDFs for each floor of each residential building; 3) Monte Carlo sampling: probabilistic sampling from PDFs with sample size proportional to population distribution; 4) aggregation of probabilistic samples on the municipality scale. Note: PDF, probability distribution function.

Missing data on the number of floor levels was imputed by using information on the building type, building age, and number of households. If data on building type, building age, and number of households was available, then a multivariate linear model (R package *stats*, function *lm*[16]) using the aforementioned information as predictors and number of floor levels as target was built with all data from the same chunk (see the section "Computational aspects"), that is, a local model was built. The rounded predicted value of floor level was used for imputation. In a few cases, no data on the building type and building age were available. In these cases, the number of floor levels was estimated using a linear model, with the number of households as single predictor and the rounded predicted value, was used for imputation. Missing data for building age was not imputed because in the survey data a class with not-applicable ("NA") values for building age existed to which these cases were assigned.

### Modeling Approach

The modeling approach (Figure 3) was based on a multistage procedure:

1. Model building: The harmonized data from the national survey and relevant predictor data (Figure 2) were used to train a regression model that was able to predict conditional quantiles of indoor radon levels on the dwelling scale.
2. Prediction: The model was applied to make a prediction of conditional quantiles for each floor level of each residential building in Germany. Based on the quantile predictions and assuming lognormality, a probability density function (PDF) was fitted for each floor level of each residential building.
3. Monte Carlo sampling: Probabilistic samples were drawn from each PDF. The sample size was proportional to the number of inhabitants. Thus, different PDFs were combined and population weighted.
4. Aggregation: The data of generated probabilistic samples in step 3 were aggregated on the administrative level of interest (national, federal states, districts, municipality) to calculate the desired statistical characteristics, such as the arithmetic mean (AM) or the probability of households exceeding a concentration of $300\ \mathrm{Bq/m^3}$.

***Model building.*** The model applied for this analysis was the quantile regression forest (QRF) model, which is a machine learning algorithm. It is a variation of the well-established random forest algorithm. The random forest was introduced by Breiman[66] and represents an ensemble of regression or classification trees. In this study, the implementation in the R package *partykit*[67] was used. In contrast to the classical implementation by Breiman, the function *cforest* in *partykit* builds conditional inference trees[68] using statistical test procedures both for predictor selection at the splits and for the definition of stopping criteria.[69]

QRFs were introduced by Meinshausen,[70] who showed that the random forest model can be used to predict not only the conditional mean, but also the entire conditional distribution. By estimating the distribution from an ensemble of realizations, the QRF model follows the same approach as conditional simulation in classical geostatistics.

Model building generally follows the approach described by Petermann et al.[71] In short, 15 candidate predictors were spatially joined with locations of indoor radon observations. The loss function was the root mean squared error (RMSE). Because the target variable, indoor radon concentration, has a skewed (near lognormal) distribution and was not transformed before model training, high observations were implicitly given more weight, that is, high observations had a greater leverage on model fit.

Predictors were selected by forward feature selection (R package *CAST*[72]) via a spatial 10-fold cross-validation approach. The size of spatial blocks (squares in our case) was set to 40 km (R package *blockCV*[73]). The motivation for predictor selection was to find only those predictors that are informative and thus contribute to improving the predictive performance of the model. The final set of selected predictors consisted of eight environmental predictors [*a*) radon in soil, *b*) outdoor temperature, *c*) slope, *d*) soil permeability, *e*) soil moisture, *f*) precipitation, *g*) wind exposure, and *h*) tectonic faults] (Figure 2) and four building-related predictors [*a*) floor level, *b*) building age, *c*) building type, and *d*) number of households]. All environmental predictors were continuous (Figure 2). Building-related predictors were categorical: floor level (basement; souterrain; ground floor, first floor, second floor, and third floor or higher), building age (<1945, 1945–1980, 1981–1995, 1996–2005, and >2006), building type (single- and two-family house, townhouse/row house and semi-detached house, multifamily house, apartment building, high-rise apartment building, terrace house, farm house, and office building), and number of living units (1, 2, 3–6, 7–12, and >12).

The hyperparameter *mtry* (number of predictors randomly selected and evaluated at each split) was tuned, and an optimal value of 4 was found. Then, the final model with selected predictors and $mtry = 4$ was fitted with 500 regression trees (*ntree*) using all available indoor radon observations.

The importance of the individual predictors in the model was estimated by a permutation-based approach (R package *vip*[74]). The predictive model is available as an interactive web application at https://model.radonmap.info.

***Prediction.*** The QRF model was used to make predictions of nine percentiles (10th, 25th, 50th, 75th, 80th, 85th, 90th, 95th, and 98th) of indoor radon concentration for each floor level of each residential building in Germany. The R scripts used for model building and prediction described in the section "Modeling Approach" can be found at https://github.com/EricPetermann/scripts_for_IRC_prediction_Germany. We created a browser-based web application (https://model.radonmap.info) for conducting predictions at the dwelling scale for user-defined locations and building characteristics.

A distribution function (3-parameter lognormal) was fitted (R package *rriskDistributions*[75]; function *get.lnorm.par*) using data of the predicted quantiles to estimate a PDF for each prediction. The three-parameter lognormal distribution function is very similar to an ordinary lognormal distribution, but it has a location shift to account for the local background values of the quantity of interest. The reason for using a three-parameter lognormal distribution was to avoid indoor radon predictions being lower than the local outdoor radon estimate, which would be implausible from a physical point of view.

To achieve this goal, the local estimate of the outdoor radon concentration (Figure 2; Supplemental Material, "S1. Environmental predictor data") was assigned to the predictions by a spatial join. To derive the distribution parameters, the outdoor radon value was subtracted from the predicted indoor radon quantiles, then an ordinary lognormal function was fitted, and finally the outdoor radon value was added after probabilistic sampling. During the fitting process, the higher percentiles were weighted more heavily to optimize the fit in this range. As a result, for each floor level of each residential building, three parameters were determined to describe the PDF: *a*) *meanlog*, *b*) *sdlog*, and *c*) *offset* (outdoor radon).

***Monte Carlo sampling.*** Monte Carlo sampling was conducted using the PDF estimated for each floor level (see the section "Prediction") using the R package *stats*; function *rlnorm*. Random samples were drawn from each PDF for propagation of prediction uncertainty, as well as for combining and weighting individual predictions. The sample size was set to be proportional to the population size, that is, the expected number of inhabitants per floor level was used to weight the individual floor-level based predictions. The sample size was determined by multiplying the expected number of inhabitants per floor level by a factor of 10 and rounded afterward to guarantee a meaningful representation of all floor levels. This was done because the sample size must be a positive integer and the expected number of inhabitants in many basements is <1. By this procedure 840,224,604 Monte Carlo samples were produced—a factor of 10 higher than the actual population size (Figure S2).

***Aggregation of results.*** A key identifier [Amtlicher Gemeindeschlüssel (AGS value)] was assigned to each random sample for efficient post-processing. This key allows unequivocal attribution of a random sample to a municipality (*Gemeinde*), district (*Landkreise und kreisfreie Städte*), and federal state (*Bundesländer*). Samples were grouped according to the key identifier and desired statistical parameters were calculated: AM; arithmetic standard deviation (SD); geometric mean (GM); geometric SD (GSD); 50th, 90th, 95th, and 99th percentiles; and exceedance probabilities of 100, 300, 600, and $1{,}000\ \mathrm{Bq/m^3}$. The final set of random samples was split across several comma-separated values (CSV) files owing to memory constraints (total size of ~22 GB). Therefore, aggregate statistics were calculated using the R packages *arrow*[76] and *dplyr*,[77] which allow querying and processing the data without the need of reading data into local memory.

***Computational aspects.*** The prediction was done for a large dataset comprising ~21.3 million residential buildings, which have, on average, 2.9 aboveground floors. In addition, a prediction was also conducted for the basement. Thus, quantile predictions for ~83 million objects were realized. This setting was required to work chunk-wise, that is, working on only a small subset of the data at one time to optimize computational speed and stability for reading, geoprocessing, predicting, distribution fitting, Monte Carlo sampling, and writing of data. The optimal (i.e., fastest overall progress) setting for the chunk size in our case was to process 5,000 buildings per chunk, which resulted in ~4,400 chunks. The chunk-wise computation allowed working in parallel and to distribute computation across several cores. The entire workflow of the predictive task took ~3,000 CPU h. The R scripts used for data processing described in the section "Modeling Approach" can be found at https://github.com/EricPetermann/scripts_for_IRC_prediction_Germany. All data were analyzed using R (version 4.2.2; R Development Core Team).

## Results

### *Survey Results*

The household survey results ($n = 14{,}053$ measurements, from 7,061 households) for indoor radon sampling showed an AM ± SD of $78 \pm 126\ \mathrm{Bq/m^3}$, a GM ± GSD of $49 \pm 2.32\ \mathrm{Bq/m^3}$, a 50th percentile of $44\ \mathrm{Bq/m^3}$, a 90th percentile of $151\ \mathrm{Bq/m^3}$, and a 95th percentile of $241\ \mathrm{Bq/m^3}$ for Germany. Minimum and maximum values were 10 and $2{,}000\ \mathrm{Bq/m^3}$, respectively. The exceedance frequencies of the 100-, 300-, 600-, and 1,000-$\mathrm{Bq/m^3}$ thresholds were 18%, 3.5%, 0.65%, and 0.35% of the individual measurements, respectively. Although the spatial distribution of the survey samples was roughly proportional to the general population distribution, a deviation from the population characteristics was observed as a result of sampling bias.

***Representativeness regarding floor level.*** Figure 4 presents the floor distribution of the population in Germany based on the population and household data provided by the BKG[65] (hereafter referred to as BKG data) compared with the distribution of samples in the indoor radon survey. BKG data[65] lack information on basement and souterrain occupancy. Nevertheless, it can be stated that up to 35.1% of the population live on the second floor or higher, whereas up to 34.8% of the population live on the ground floor based on BKG data (Figure 4A; Excel Table S3). In contrast, the floor-level distribution in the sampled data (Figure 4B) deviates from the population characteristics as follows: 8.1% of the samples ($n = 1{,}150$) are from second floor or higher, 46.8% ($n = 6{,}583$) are from the ground floor. These numbers clearly highlight the tendency of disproportionately high numbers of samples to be from lower floor levels. Further, 14.5% of the total samples were from basement floors, also indicating a significant overrepresentation of this floor.

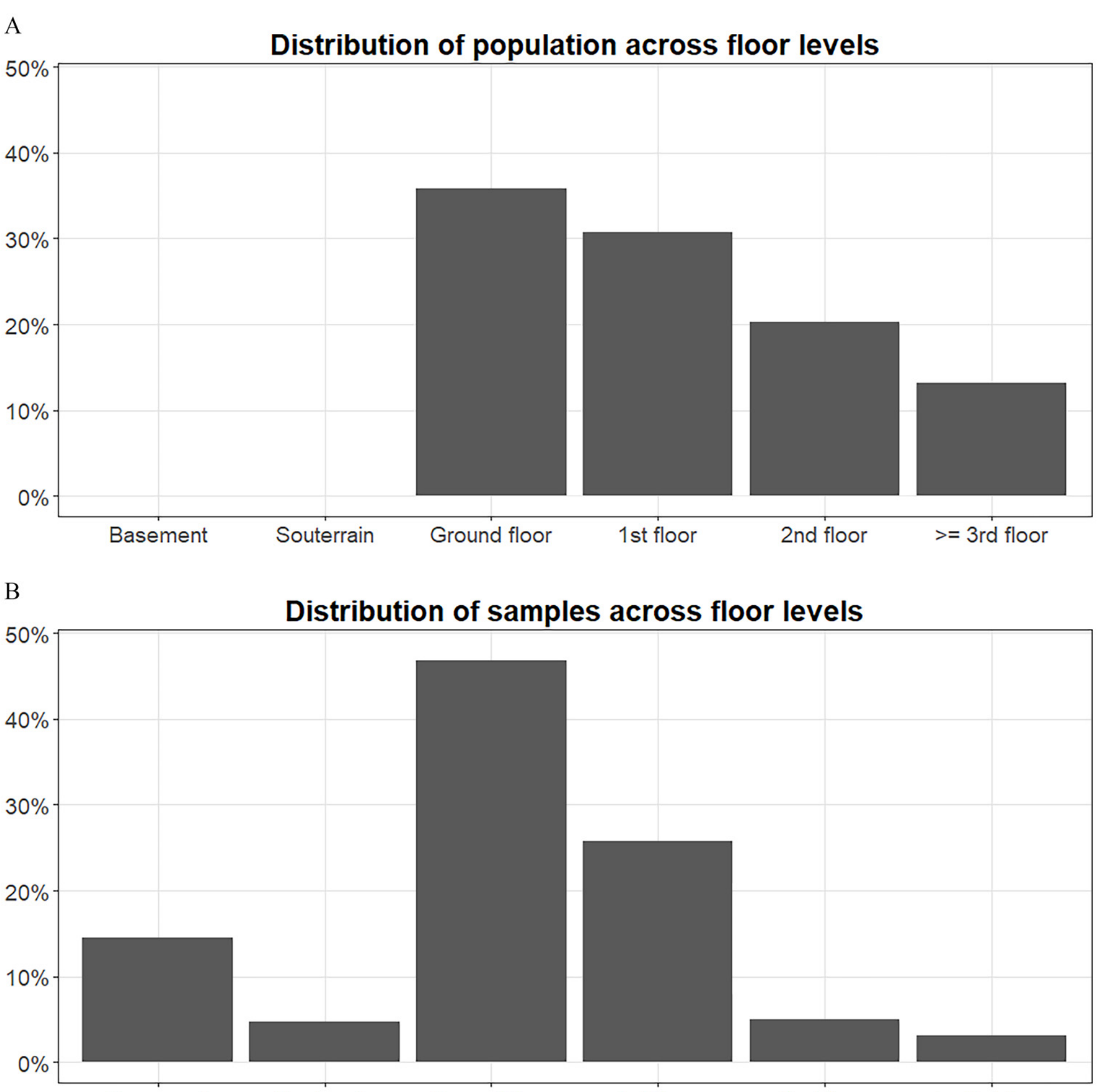

**Figure 4.** Comparison of sample and population characteristics with respect to floor-level distribution, in the national indoor radon survey in Germany, 2019–2021. (A) Distribution of population across floor levels is based on residential building stock data (no basement information available; $n = 21.9$ million); (B) distribution of samples across floor levels is based on data provided by participants via questionnaire for each measurement ($n = 14{,}053$). Data are available in Excel Table S3.

***Representativeness regarding radon in soil.*** Another important factor in evaluating the representativeness of the indoor measurements is their distribution with respect to the geogenic radon source. Thus, the distribution of the population with respect to the radon concentration in soil was compared with the distribution of the samples with respect to the radon concentration in soil. For this purpose, the data of population distribution in Germany and sample locations in the indoor radon survey were each spatially linked with the map of radon concentration in soil (Figure 2). The resulting distribution is shown as a probability plot in Figure 5 (with data shown in Excel Table S4). If the sample locations in the survey were fully representative of spatial distribution of the German population with respect to radon concentration in soil, the two curves would be congruent. Indeed, the population and sample distributions are very similar, at least for percentiles 1 to ∼90.

### *QRF Model*

***Predictor importance.*** The two top predictors of indoor radon concentration are floor level and radon concentration in soil, followed by building age, outdoor temperature, and slope (Figure 6). A considerable importance has also been found for soil gas permeability, soil moisture, precipitation, number of living units, wind exposure, and building type. The importance of the predictor tectonic faults is negligible in our case. Overall, environmental- and building-related predictors were roughly equally important. Information on the effect of individual predictor levels can be derived from the partial dependence plots (Figure S3).

***Predictive performance.*** The performance of the predictive model—tested by a 10-fold spatial cross-validation—was analyzed with respect to the conditional mean, conditional quantiles, and prediction intervals. Predictive accuracy for the conditional mean was suboptimal, as evidenced by a RMSE of 110.5 and a $r^2$ (coefficient of determination) of 0.24. Similarly to previous studies,[35,71,78,79] an underestimation of high values and overestimation of low values was observed, that is, there was a general smoothing tendency of the predictions (Figure 7).

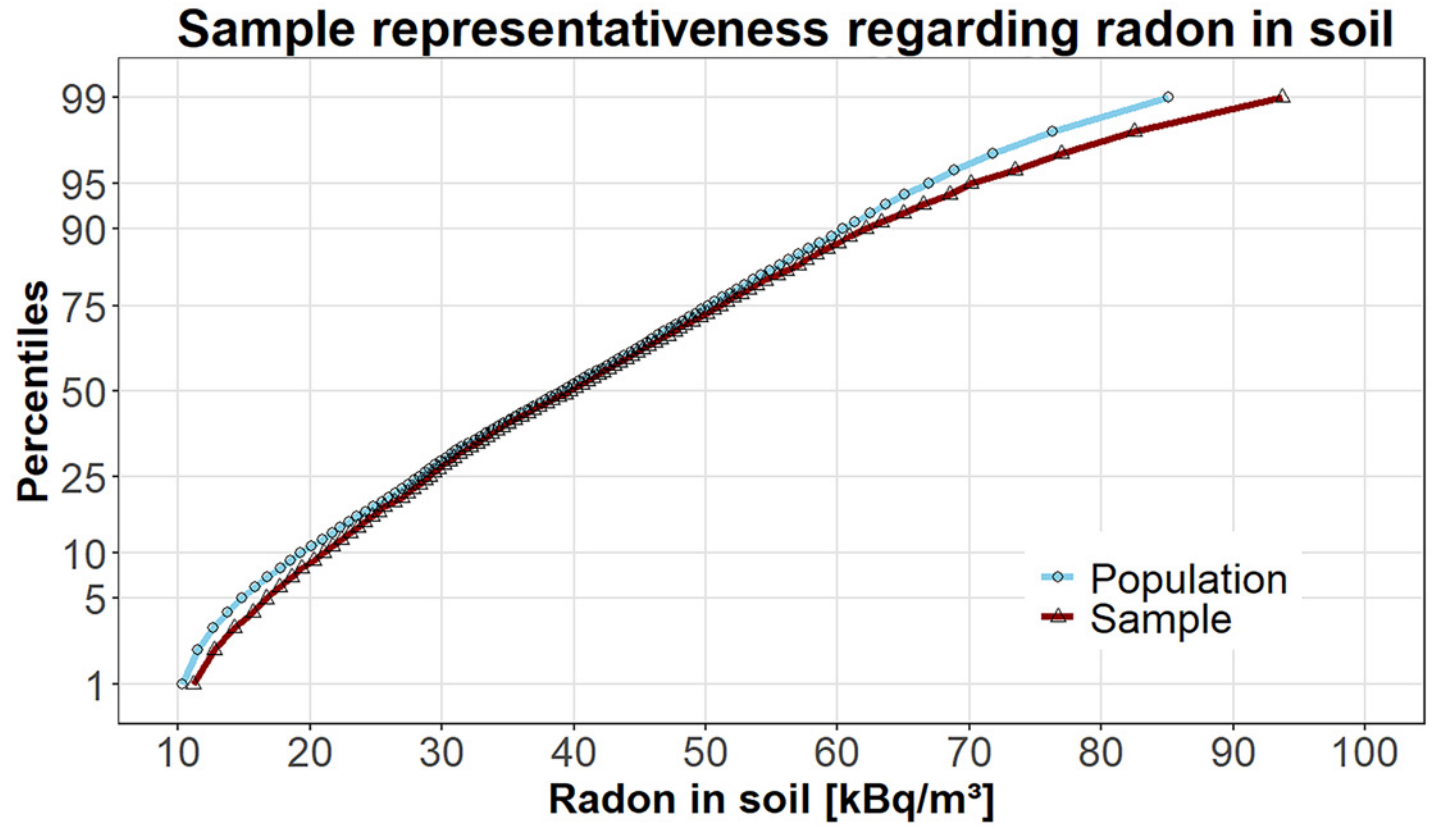

**Figure 5.** Comparison of sample and population characteristics with respect to radon concentration in soil (see Figure 2). Values for "sample" are based on the location of individual measurements in the indoor radon survey ($n = 14,053$) and values for "population" ($n = 83$ million) are based on the population distribution in Germany in 2022 (source: BKG). Data are available in Excel Table S4.

In Figure 7 "observed" refers to measured data, and "predictions" to modeled values for those locations for which measured data were available.

The 80% (10–90 percentile) and 50% (25–75 percentile) prediction intervals cover 78.2% and 49.1% of test observations, respectively, which is very close to the desired values. Consequently, a good overall performance can be concluded. Moreover, the scatter of observations outside the prediction intervals is evenly distributed over the entire range of predicted values (Figure 8). In addition, about the same proportion of the test data lies above and below the prediction interval. This observation is confirmed when focusing on specific quantiles: The quantile coverage probability (QCP) of selected quantiles is shown in Table 1 and indicates that the actual coverage is very close to the nominal coverage over the entire range of quantiles. On average, the QCP deviates from the nominal coverage by only $\sim 1\%$ point. Consequently, we can conclude that the prediction intervals are very accurate and reliable: The desired proportion of the data is covered by the respective intervals.

### *Indoor Radon Concentration*

The predicted indoor radon distribution of Germany is characterized by an AM $\pm$ SD of $63 \pm 147\,\mathrm{Bq/m^3}$ and a GM $\pm$ GSD of $41 \pm 2.27\,\mathrm{Bq/m^3}$. Selected quantiles of the 50th, 90th, and 95th percentiles are 36, 115, and $180\,\mathrm{Bq/m^3}$, respectively. The exceedance frequencies for 100, 300, 600, and $1,000\,\mathrm{Bq/m^3}$ are 12.5% (10.5 million people), 2.2% (1.9 million people), 0.67% (560,000 people), and 0.25% (210,000 people), respectively.

Figure 9 presents the estimated quantities of indoor radon concentration at different spatial resolutions: federal states, districts and municipalities. The same data are also available on an interactive website (https://indoor.radonmap.info). The AM and the exceedance probability of $300\,\mathrm{Bq/m^3}$ ($>300\,\mathrm{Bq/m^3}$) reveal similar spatial patterns. The most affected federal states are Saxony [AM $= 100\,\mathrm{Bq/m^3}$; probability $>300\,\mathrm{Bq/m^3}$ (P300): 5.9%], Thuringia (AM $= 103\,\mathrm{Bq/m^3}$; P300: 5.6%), and Bavaria (AM $= 85\,\mathrm{Bq/m^3}$; P300: 4.0%). However, when focusing on the higher spatial resolution of districts or municipalities, it becomes clear that affected areas also exist in Baden-Württemberg, Rhineland-Palatinate, Hesse, North Rhine-Westphalia, Saxony-Anhalt, Mecklenburg-Western Pomerania, Lower Saxony, and Schleswig-Holstein (see top-left plot in Figure 9 for location of the federal states). At the district level, 18 of 401 districts exceeded an AM of $150\,\mathrm{Bq/m^3}$, as well as a 10% probability of exceeding the $300\,\mathrm{Bq/m^3}$ reference level. At the municipality level, $\sim 900$ of 10,885 municipalities exceeded an AM of $150\,\mathrm{Bq/m^3}$, as well as a 10% probability of exceeding the $300\,\mathrm{Bq/m^3}$ reference level. In most cases, the areas with the highest AM and P300 values were located in low mountain ranges (e.g., Erzgebirge: Saxony, Fichtelgebirge: Bavaria, Harz: Saxony-Anhalt, Lower Saxony; Black Forest: Baden-Württemberg) or the Alps (Bavaria) or in areas with moraine deposits from the most recent glacial advance (Mecklenburg-Western Pomerania, Schleswig-Holstein), which have a higher geogenic radon potential.[71]

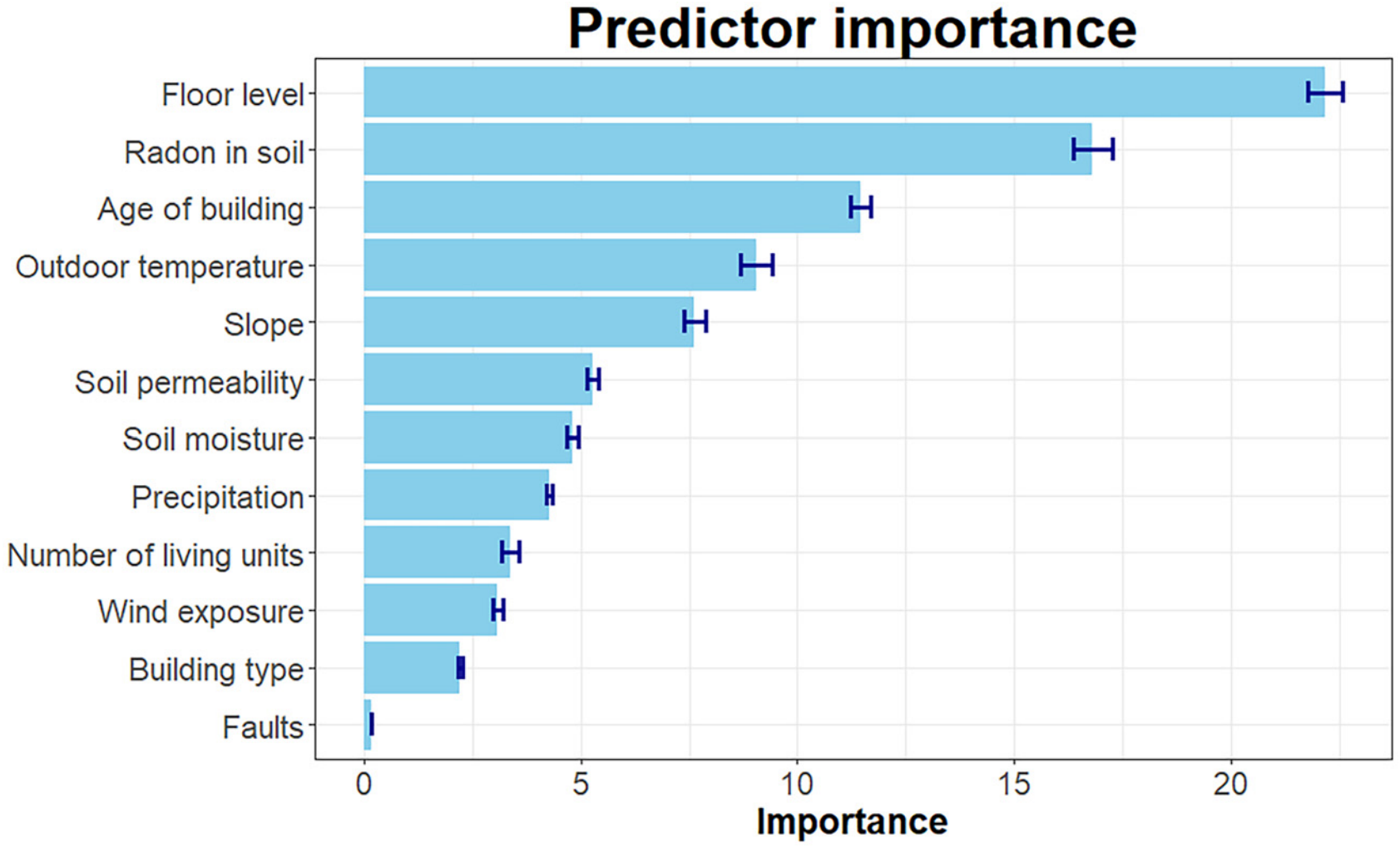

**Figure 6.** Ranking of predictors by importance for indoor radon prediction with the quantile regression forest model. The importance was measured by calculating the increase of the model's prediction error after permutation (i.e., randomly shuffling values) of the predictor of interest. Data are available in Excel Table S5.

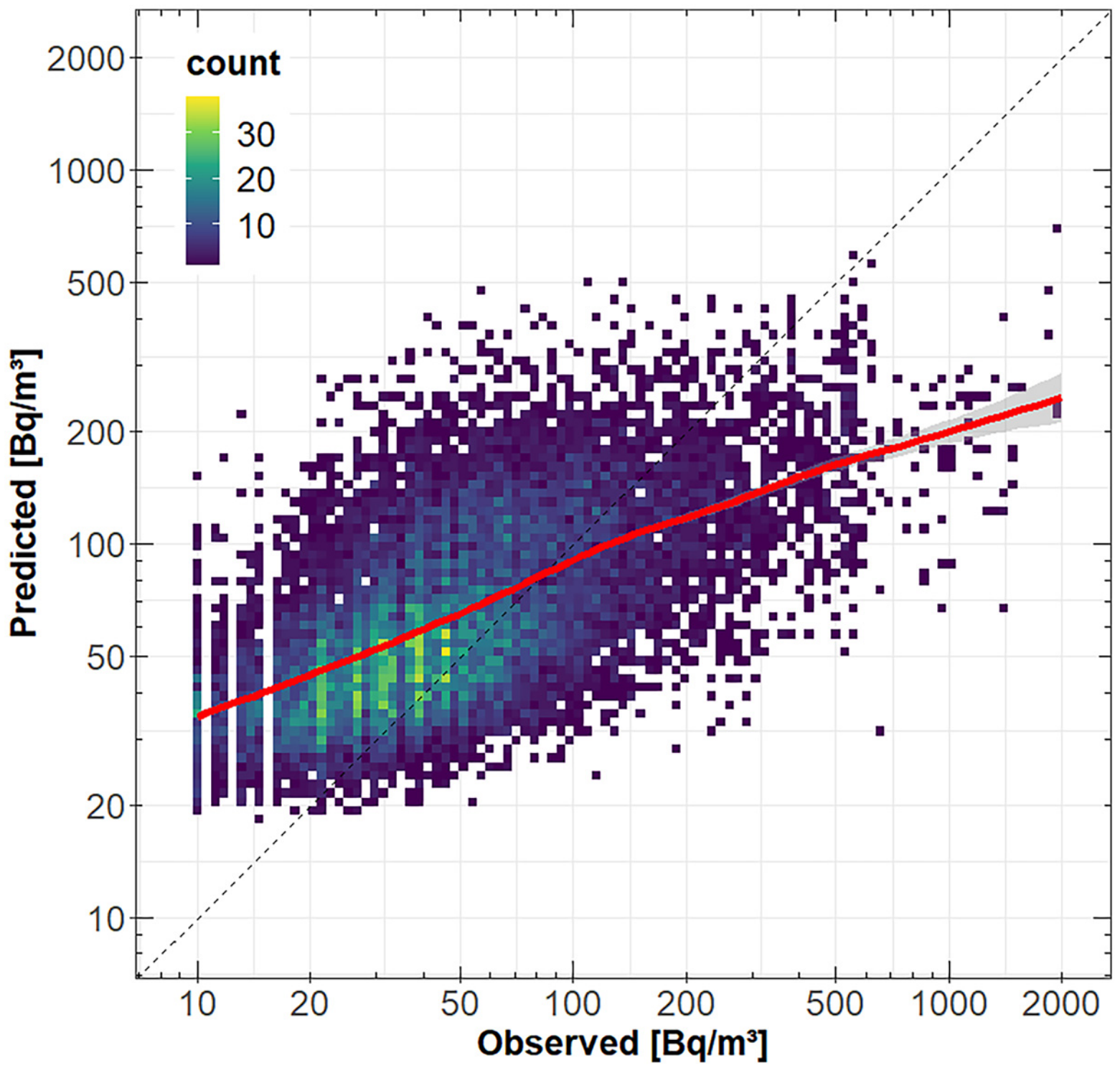

**Figure 7.** Model performance of quantile regression forest model for predicting the conditional mean of indoor radon levels. Axes are on a log scale. The red line shows the smoothed conditional mean estimated by a generalized additive model. Results are based on 10-fold spatial cross-validation ($n = 14,053$). Data are available in Excel Table S6.

The quantities of AM and P300 are indicators of the individual risk, whereas the number of people exposed to indoor radon $>300\,Bq/m^3$ (third row of Figure 9) represents an indicator of the collective risk. As a consequence of population density, the highest collective indoor radon risk can be found in big cities, such as Munich, Berlin, Cologne, and Hamburg, even though the geogenic radon hazard is low in most of the big cities in Germany (with the exception of Munich, which has a moderate-to-elevated geogenic radon hazard).

## Discussion

### *Modeling Approach*

The model described in this study uses the correlation between indoor radon concentration and predictor data, such as environmental conditions and building-related information. The most important predictors were identified as floor level and soil radon, which is in line with many other studies where they have been identified as key factors (see the section "Potential predictor data" for details). Floor level describes the distance to the ground, which is usually the main source of indoor radon. Radon concentration in soil in combination with the gas permeability of the soil is the key factor for geogenic radon hazard. The other top predictors also seem plausible from a physical point of view:

- Building age might reflect the quality of the buildings foundation, the building materials used, as well as the tightness of building isolation, which in turn affects air exchange.
- Outdoor temperature affects pressure differences between indoor and outdoor.
- The slope inclination of the terrain determines the contact area between building and ground.
- Precipitation (annual sum) and soil moisture (annual mean) are proxies for the local interaction of soil physical properties and climate. The related processes such as radon emanation and radon transport in the soil affect the geogenic radon potential.
- Wind exposure index reflects the relative terrain position (e.g., less exposed in valleys, more exposed on hilltops), which acts as a proxy for average wind speed, which can affect the indoor/outdoor pressure differences and, thus, radon entry.

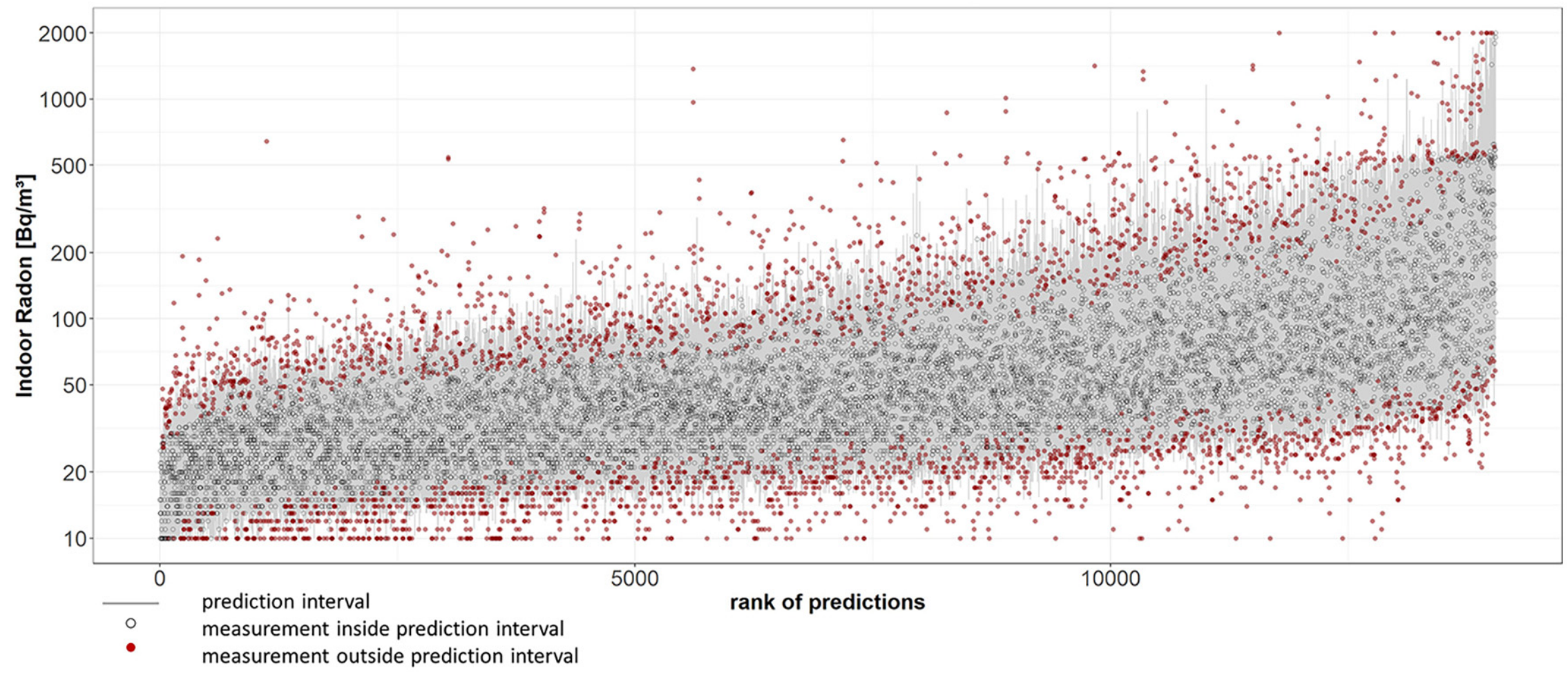

**Figure 8.** Prediction intervals (10–90 percentiles) of the quantile regression forest model vs. observed values of indoor radon levels. Predictions are shown for test data from a 10-fold spatial cross-validation. Data are sorted by increasing estimates of the conditional median. Prediction intervals are visualized as gray vertical bars. Measurements inside the prediction interval are depicted as black open circles, predictions outside the prediction interval are depicted as dark red circles. Data are available in Excel Table S7.

**Table 1.** Quantile coverage probability of predicted quantiles for the quantile regression forest model predicting indoor radon levels in Germany.

| Quantile (nominal) (percentile) | Quantile coverage probability (actual) (%) |
|---|---|
| 10th | 11.0 |
| 25th | 26.3 |
| 50th | 50.9 |
| 75th | 74.6 |
| 80th | 79.6 |
| 85th | 84.5 |
| 90th | 89.0 |
| 95th | 93.9 |
| 98th | 96.9 |

The availability of point-scale building-related data and high-resolution environmental data (maximum grid cell size of 500 m) allowed characterization of indoor radon distribution even at the municipality level. Consequently, the applied modeling approach was able to *a*) correct for sampling bias, and *b*) provide predictions at the subnational level. The analysis of model performance revealed that accurate prediction of point estimates, such as the conditional mean of indoor radon, is not yet possible, which is consistent with many other studies.[31,35,80,81] This finding is reflected in a large local prediction uncertainty and the resulting wide prediction intervals. However, as demonstrated in this study, a powerful model, such as the QRF, combined with currently available predictor data, allows for accurate characterization of the prediction uncertainty. The QRF model is increasingly applied in recent years. Many studies with varying objectives, such as digital soil mapping,[82] heat wave prediction,[83] or spatial prediction of groundwater levels,[84] have proven the ability of the QRF model to provide robust estimates of prediction intervals. The ability to provide reliable prediction intervals, in turn, allowed an effective propagation of the predictive uncertainty into indoor radon concentration variability. The latter was achieved by combining the QRF-based predictions of the conditional distributions (PDF for each floor in each residential building) with a Monte Carlo sampling procedure. The use of high-resolution population distribution data allows population weighting of the predictions by making the sample size in the Monte Carlo sampling procedure proportional to the expected number of people per floor.

### *Uncertainty and Limitations*

The sample size of the indoor radon survey was set considering the requirement to allow a robust characterization of the statistical moments at the national level. However, the sample contained *a*) too many measurements in the basement and on the ground floor, and *b*) too many measurements from areas with high radon in soil concentration (Figure 5; Excel Table S4). The latter could be a consequence of deviations from the desired sample size in some districts (see the section "Survey Results") or could be a result of partially voluntary participation in the survey, or both. In consequence *a*) an over-sampling of rooms located in more exposed basements and an under-sampling of rooms located in less exposed higher floors, and *b*) an overrepresentation of buildings in areas with higher radon concentration in soil can be detected. Thus, descriptive statistics of the survey are likely to overestimate the (unknown) true concentration at the national level. This sampling bias requires the use of a model-based approach to provide unbiased estimates of indoor radon concentration. In addition, the sample size was too small (intentionally; owing to budget constraints) to allow characterization of indoor radon distribution at the subnational level (federal states, districts, municipalities) based on descriptive statistics alone.

There are several sources of uncertainty related to the applied modeling approach. The main reason for the large prediction uncertainty is likely due to missing relevant prediction data, such as building characteristics and behavior of the residents (e.g., air tightness of the building, ventilation intensity, heating system, window opening), as well as the possible inaccuracy of the existing predictor data (e.g., insufficient map resolution, suboptimal classification). Probably the largest source of uncertainty is the lack of spatially resolved data on the occurrence and usage of basements. This information is important because in most buildings, radon concentration is greatest in basements.[14,22,85] In this study, we parameterized basement occupancy by an "educated guess" (see the section "Floor-level distribution of the population" and Figure S2) of 30% basement occupancy relative to the occupancy of each upper floor for single- and two-family houses, row houses, for example, and 5% for multifamily houses, apartment blocks, for example.

Model performance, such as the reliability of prediction intervals (see the section "Predictive performance"), is evaluated at the global scale. Consequently, local deviations from the predicted concentration distribution may occur if the predictor maps are not accurate in a given area or if radon-relevant conditions exist at the local scale that are not reflected by the predictors [e.g., in (post)-mining areas]. Because the predictor maps are also the result of predictive modeling (interpolation), they are also subject to uncertainty. The more important a single predictor is, the more sensitive the final prediction becomes to the uncertainty of that predictor in that area. The most important environmental predictor (Figure 4) is soil radon concentration (see the Supplemental Material, "S1. Environmental predictor data" for more details on mapping). In turn, an important predictor of the soil radon map is the geologic map, that is, the classification system and spatial extent of a single geologic unit influence the soil radon map. Therefore, municipality-scale predictions should be interpreted with consideration of the predictors used in the model (Figure 3) and their respective values in the predictor maps (Figure 2) at the local level and consideration of possible local specifications. In general, uncertainty is expected to be significantly lower on the national scale but increased at higher spatial resolutions.[86] In our case, uncertainty at the national scale is expected to be driven by uncertainty of basement prevalence and occupation, whereas uncertainty at the district or municipality scale is expected to be driven by both uncertainty of basement prevalence and occupation, as well as predictor data uncertainty.

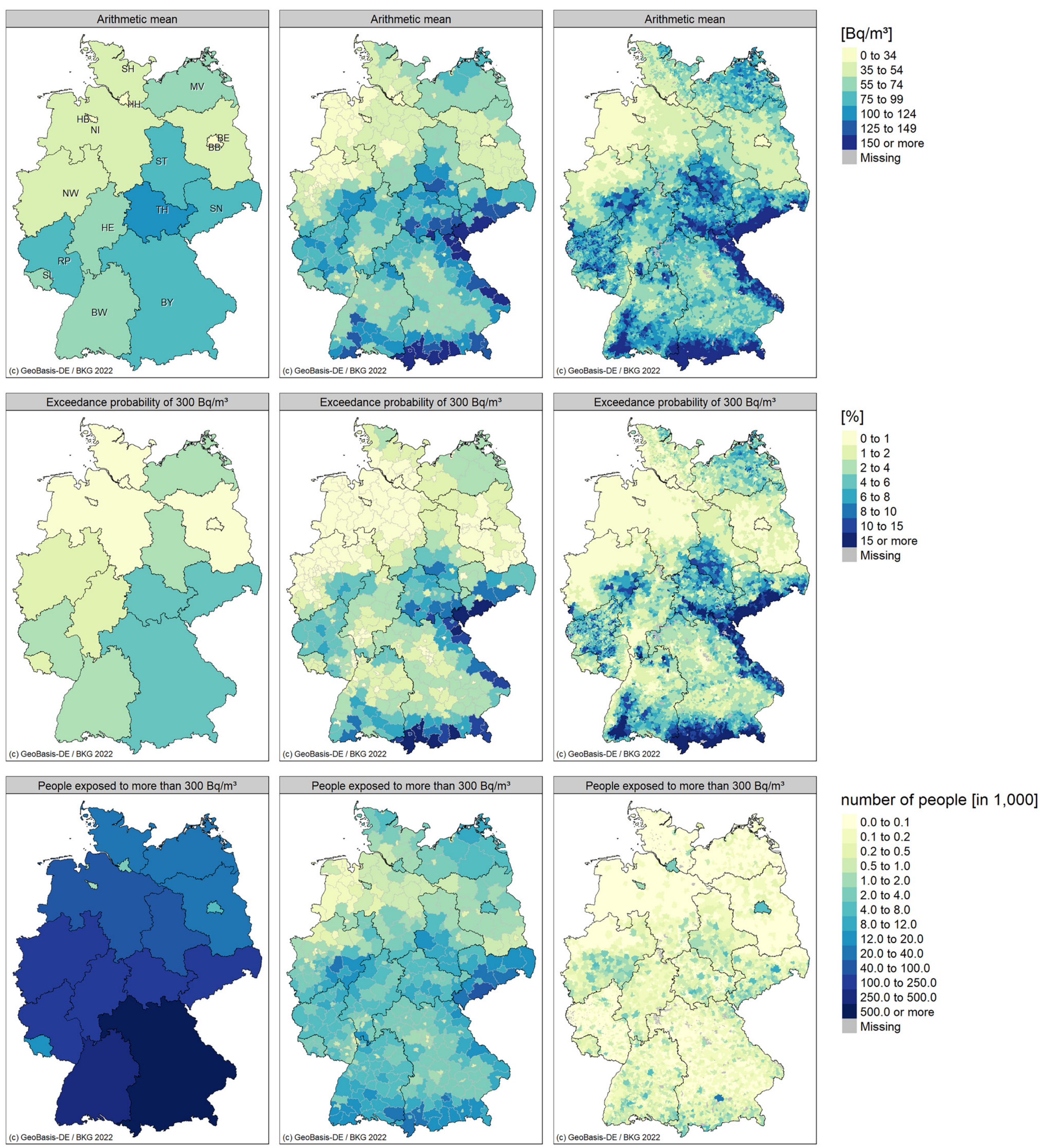

**Figure 9.** Maps of indoor radon characteristics in Germany, estimated with a model-based approach, at the administrative level of federal states (*Bundesländer*), districts (*Landkreise und kreisfreie Städte*), and municipalities (*Gemeinden*) in the first, second, and third columns, respectively. The depicted quantities are the AM, relative exceedance of $300\,\mathrm{Bq/m^3}$ (%), and the number of people who were exposed to $>300\,\mathrm{Bq/m^3}$ in the first, second, and third row. On the municipality level, predictions are provided only for municipalities with at least 100 inhabitants (i.e., 1,000 Monte Carlo samples). Interactive maps can be found at https://indoor.radonmap.info. Data are available in Excel Tables S8–S10. Note: AM, arithmetic mean; BB, Brandenburg; BE, Berlin; BW, Baden-Württemberg; BY, Bavaria; HB, Bremen; HE, Hesse; HH, Hamburg; MV, Mecklenburg-Western Pomerania; NI, Lower Saxony; NW, North Rhine-Westphalia; RP, Rhineland-Palatinate; SH, Schleswig-Holstein; SL, Saarland; SN, Saxony; ST, Saxony-Anhalt; TH, Thuringia.

In general, the annual mean of indoor radon is considered to be a reliable estimate of the long-term mean value. However, several studies (see the review by Antignani et al.[87]) have shown that there are interannual variations at the building level. For larger areas under investigation or multiannual surveys, these year-to-year variations are likely to average out to some extent and to be of less relevance. However, because our survey was conducted in the period 2019–2021, the resulting estimates, in a strict sense, refer to only that period. The annual average of the study period could deviate from the long-term mean if there were significant large-scale changes in relevant factors, such as climate (e.g., extreme years in terms of temperature or precipitation), lifestyles, or national emergencies, that affect ventilation rates (e.g., COVID pandemic in 2020/2021), were present. Furthermore, beyond estimating the distribution of indoor radon concentrations, changing lifestyles (e.g., remote working, lockdown) also affect the occupancy of houses/apartments, which affects concentration at the individual level in terms of occupancy times.

Uncertainty at the national level can be further reduced by implementing information on the prevalence of basements and their occupancy (regionalized and differentiated by building type). Uncertainty of predictions for individual building and floors could be further reduced if spatially exhaustive data on the building material, the ventilation system and the energy efficiency status of a building (e.g., building codes, energy retrofitting) becomes available.

In summary, the prediction uncertainty is large but can be described accurately by the estimated prediction intervals. Therefore, it can be assumed that propagation of the prediction uncertainty via a probabilistic procedure, such as Monte Carlo sampling, provides a reasonable approximation to the true variability of individual radon concentration.

### *Indoor Radon Distribution and Radon Policy*

The results from the present study reveal a higher mean indoor radon concentration in Germany ($AM = 63\ Bq/m^3$ and $GM = 41\ Bq/m^3$ indoor radon concentration) than previously estimated by Menzler et al.,[88] who determined an AM of $49\ Bq/m^3$ and a GM of $37\ Bq/m^3$. The main reason for this difference is likely to be the consideration of basement occupancy—a key factor deliberately neglected by Menzler et al.[88] owing to the lack of data at the time. The differences in the spatial patterns of indoor radon between the two studies are primarily due to *a*) the availability of more and better-resolved predictor data nowadays, especially in terms of soil radon concentration and building information; and *b*) the predictive modeling approach that fully reflects the distribution of radon-relevant factors in the entire population. It should be noted that possible temporal changes cannot be readily inferred from the comparison of the two studies owing to methodological differences and assumptions inherent in the studies.

The presented maps of the AM and the probability of the exceedance of the reference level of $300\ Bq/m^3$ represent the individual risk given that they combine information on hazard (environmental conditions) and vulnerability (housing characteristics). The map of the total number of people exceeding the reference level, on the other hand, is a measure of collective risk. The spatial patterns are generally consistent with the maps of buildings affected by radon (i.e., expected value $>300\ Bq/m^3$ on the ground floor) produced by Petermann and Bossew[33] using a much simpler approach that built solely on geogenic radon potential as single predictor. However, noticeable differences can be observed in some areas, especially in the south and west of Germany. The number of radon-affected buildings was estimated as 350,000,[33] which is a factor of 5 to 6 lower than the number of radon-affected people (with exposures $>300\ Bq/m^3$), which was estimated as 1.9 million people in this study. These figures seem plausible because, in most cases, several people are affected by elevated radon concentrations if a value of $>300\ Bq/m^3$ is expected on the ground floor of a building. As a rule, however, not all persons living in this building are affected in every case, especially on the upper floors. On the other hand, people may also be affected by an exceedance of the reference value if they stay in the basement, even if the ground floor is not affected, and the building would not be classified as radon-affected.

Another interesting feature is that large cities usually have lower values (i.e., AM, P300) than the surrounding areas—and this even in cases where the city and the surrounding area have a comparable geogenic radon risk. The reason for this difference in indoor radon concentration is that the population density in urban areas is much higher than in their surrounding areas. As a result, multifamily buildings and apartment buildings are much more common in urban areas, which means that the proportion of people living in upper floors is also greater in urban areas than in rural municipalities. In other words: In an area with the same geogenic radon hazard, indoor radon concentration would be higher in rural areas than in urban areas because more people in rural areas—owing to differences in settlement patterns—live in lower floors and thus closer to the geogenic radon source. For the sake of health protection, radon priority areas (RPAs) must be defined from 2020 onwards according to Euratom Basic Safety Standards (EU-BSS),[89] a Euratom directive. In Germany, RPAs are delineated by the competent authorities of the federal states based on various information sources, such as national radon maps or relevant local information. The formal criterion is a 10% exceedance probability of the reference level ($300\ Bq/m^3$) in a residential building for a standard situation "ground floor of a residential building with basement" on at least 75% of the area of the respective administrative unit. Currently, $\sim 2\%$ of the area of Germany (inhabited by 1% of the population) is designated as an RPA. We found that only 9% of the population likely to be exposed to indoor radon concentrations $>300\ Bq/m^3$ lives in an RPA. This result is consistent with our estimate of 7% of radon-exposed buildings (i.e., expected exceedance of $300\ Bq/m^3$ on the ground floor) in a previous study.[90]

It is important to note, the exceedance probabilities presented in this study, which focus on the average concentration of the population, are not directly comparable with the reference building defined for the delineation of the RPA. However, when analyzing the RPA status (yes/no) of the 10 districts with the highest indoor radon concentration predictions, only 4 districts are fully or partially designated as RPAs—in 6 districts not a single municipality is a designated RPA. As far as municipalities are concerned, as shown in the section "Indoor Radon Concentration," in $\sim 900$ cases the exceedance probability of $300\ Bq/m^3$ (reference value) is $>10\%$ and the AM is $>150\ Bq/m^3$. However, in Germany a total of only 210 municipalities are designated as RPAs. On one hand, the fact that only $\sim 1\%$ of the population live in RPAs, but $\sim 9\%$ of cases of reference value exceedances are to be expected in RPAs show the accuracy of the RPAs delineated due to the significantly higher level of affectedness. On the other hand, it becomes obvious that the vast majority of the German population ($\sim 90\%$) exposed to indoor radon levels $>300\ Bq/m^3$ is expected to live outside RPAs. We therefore recommend that these figures should be considered for policy measures in the near future to optimize radiation protection of the population.

## Conclusion

A new estimate of indoor radon distribution for Germany was produced, providing information on relevant statistical characteristics such as the AM or the exceedance probability of $300\ Bq/m^3$ at different administrative levels (country, federal state, district, municipality). Thus, updated indoor radon maps for Germany are provided (https://indoor.radonmap.info), and the prediction model (https://model.radonmap.info) is available on interactive websites that facilitate interaction with the results and is intended to increase the comprehensibility of the model.

The application of a QRF model that was trained with measurement data from a recent harmonized national indoor radon survey and high-resolution environmental and building-related predictor data provided a reliable characterization of prediction intervals and, thus, estimation of probability density functions for each floor level of each residential building in Germany. The results of this study could be used for epidemiological studies for such as calculating the proportion of lung cancer cases attributable to radon under consideration of additional data, such as smoking behavior. Furthermore, the model could also be used in case–control studies for predicting indoor radon concentrations for cases for which no measured data are available. The presented modeling approach can also be applied to other countries or regions given that a comprehensive set of indoor radon measurement data, relevant predictor data, and an exhaustive dataset on the residential building stock, including relevant building characteristics, are available.

## Acknowledgments

Author contributions: survey design: J.K., V.G., P.B., and B.H.; survey realization: J.K. and V.G.; methodology: E.P. and P.B.; data analysis, modeling and software, and drafting of the manuscript: E.P.; revision of original manuscript: all authors.

Special thanks to Valentin Ziel [Federal Office for Radiation Protection (BfS)] for technical support in creating the websites. Further, we thank Felix Heinzl and Nora Fenske (both BfS) for discussion of the draft manuscript and valuable feedback. We also thank the Federal Agency for Cartography and Geodesy (BKG) for providing the dataset on buildings and inhabitants for Germany in advance. We are also grateful for the valuable input of the editorial team and three anonymous reviewers whose critical comments contributed to the quality of this paper.

The indoor radon survey was funded by the German Federal Ministry for the Environment, Nature Conservation, Nuclear Safety and Consumer Protection (project 3618S12261, *Ermittlung der aktuellen Verteilung der Radonkonzentration in deutschen Wohnungen*, to J.K.).